\documentclass{llncs}
\usepackage{listings,alltt}
\usepackage{epsf}
\usepackage{color}
\usepackage{url}
\usepackage{amstext,amsmath}
\usepackage{cleveref}
\usepackage{graphicx}
\usepackage{algorithm}
\usepackage{algorithmic}
\usepackage{tabularx}

\newcommand{\comment}[1]{}

\newcommand{\scode}[1]{{\small \texttt{#1}}}

\begin{document}

\title{Learn\&Fuzz:\\Machine Learning for Input Fuzzing}
\author{Patrice Godefroid\inst{1}
\and Hila Peleg\inst{2}\thanks{The work of this author was 
    done mostly while visiting Microsoft Research.}
\and Rishabh Singh\inst{1}}
\institute{
Microsoft Research\\ \email{\{pg,risin\}@microsoft.com} \\
\and The Technion \\ \email{hilap@cs.technion.ac.il}  \\
}

\maketitle

\newcommand{\nosample}{\texttt{NoSample}}
\newcommand{\orig}{\texttt{Sample}}
\newcommand{\fuzz}{\texttt{SampleSpace}}
\newcommand{\morefuzz}{\texttt{SampleFuzz}}
\newcommand{\origrandom}{\texttt{Sample+Random}}
\newcommand{\fuzzrandom}{\texttt{SampleSpace+Random}}
\renewcommand{\t}[1]{\texttt{#1}}
\newcommand{\quotes}[1]{``#1"}
\newcommand{\argmin}{\operatornamewithlimits{argmin}}

\begin{abstract}

Fuzzing consists of repeatedly testing an application with modified,
or fuzzed, inputs with the goal of finding security vulnerabilities in
input-parsing code. In this paper, we show how to automate the
generation of an input grammar suitable for input fuzzing using sample
inputs and neural-network-based statistical machine-learning
techniques. We present a detailed case study with a complex input
format, namely PDF, and a large complex security-critical parser for
this format, namely, the PDF parser embedded in Microsoft's new Edge
browser. We discuss (and measure) the tension between conflicting
learning and fuzzing goals: learning wants to capture the structure of
well-formed inputs, while fuzzing wants to break that structure in
order to cover unexpected code paths and find bugs. We also present a
new algorithm for this learn\&fuzz challenge which uses a learnt input
probability distribution to intelligently guide where to fuzz inputs.

\end{abstract}

%
%
%

\section{Introduction}

{\em Fuzzing} is the process of finding security vulnerabilities in
input-parsing code by repeatedly testing the parser with modified, or
{\em fuzzed}, inputs. There are three main types of fuzzing techniques
in use today: (1) {\em blackbox random} fuzzing~\cite{fuzzing-book},
(2) {\em whitebox constraint-based} fuzzing~\cite{SAGE}, and (3) {\em
grammar-based} fuzzing~\cite{purdom1972sgt,fuzzing-book}, which can be
viewed as a variant of model-based
testing~\cite{utting2006tmb}. Blackbox and whitebox fuzzing are fully
automatic, and have historically proved to be very effective at
finding security vulnerabilities in binary-format file parsers. In
contrast, grammar-based fuzzing is not fully automatic: it requires an
input grammar specifying the input format of the application under
test. This grammar is typically written by hand, and this process is
laborious, time consuming, and error-prone.  Nevertheless,
grammar-based fuzzing is the most effective fuzzing technique known
today for fuzzing applications with complex structured input formats,
like web-browsers which must take as (untrusted) inputs web-pages
including complex HTML documents and JavaScript code.

In this paper, we consider the problem of {\em automatically}
generating input grammars for grammar-based fuzzing by using
machine-learning techniques and sample inputs. Previous attempts have
used variants of traditional automata and context-free-grammar
learning algorithms (see Section~\ref{sec:related-work}).  In contrast
with prior work, this paper presents the {\em first attempt} at using
{\em neural-network-based statistical learning techniques} for this
problem. Specifically, we use {\em recurrent neural networks} for
learning a statistical input model that is also {\em generative}: it
can be used to generate new inputs based on the probability
distribution of the learnt model (see Section~\ref{sec:learning} for
an introduction to these learning techniques). We use unsupervised
learning, and our approach is fully automatic and does not require any
format-specific customization.

We present an in-depth case study for a very complex input format:
PDF. This format is so complex (see Section~\ref{pdf-struc}) that it
is described in a 1,300-pages (PDF) document~\cite{pdf-manual}. We
consider a large, complex and security-critical parser for this
format: the PDF parser embedded in Microsoft's new Edge
browser. Through a series of detailed experiments (see
Section~\ref{sec:evaluation}), we discuss the {\em learn\&fuzz
challenge}: how to learn and then generate diverse well-formed inputs
in order to maximize parser-code coverage, while still injecting
enough ill-formed input parts in order to exercise unexpected code
paths and error-handling code.

We also present a novel {\em learn\&fuzz} algorithm (in
Section~\ref{sec:learning}) which uses a learnt input probability
distribution to intelligently guide {\em where} to fuzz (statistically
well-formed) inputs. We show that this new algorithm can outperform
the other learning-based and random fuzzing algorithms considered in
this work.

The paper is organized as follows. Section~\ref{pdf-struc} presents an
overview of the PDF format, and the specific scope of this
work. Section~\ref{sec:learning} gives a brief introduction to
neural-network-based learning, and discusses how to use and adapt such
techniques for the learn\&fuzz problem. Section~\ref{sec:evaluation}
presents results of several learning and fuzzing experiments with the
Edge PDF parser. Related work is discussed in
Section~\ref{sec:related-work}. We conclude and discuss directions for
future work in Section~\ref{sec:conclusion}.

\section{The Structure of PDF Documents}\label{pdf-struc}

\begin{figure}[t]
\centering
\setlength{\tabcolsep}{12pt}
\begin{tabular}[t]{ccc}
\begin{lstlisting}
2 0 obj
<<
/Type /Pages
/Kids [ 3 0 R ]
/Count 1
>>
endobj
\end{lstlisting} &
\begin{lstlisting}
xref
0 6
0000000000 65535 f
0000000010 00000 n
0000000059 00000 n
0000000118 00000 n
0000000296 00000 n
0000000377 00000 n
0000000395 00000 n
\end{lstlisting}&
\begin{lstlisting}
trailer
<<
/Size 18
/Info 17 0 R
/Root 1 0 R
>>
startxref
3661
\end{lstlisting}\\
(a) & (b) & (c)
\end{tabular}
\caption{Excerpts of a well-formed PDF document. (a) is a sample object, (b) is a cross-reference table with one subsection, and (c) is a trailer.}\label{pdf-samples}
\end{figure}

The full specification of the PDF format is over $1,300$ pages long~\cite{pdf-manual}. Most of this specification -- roughly 70\% -- deals with the description of {\em data objects} and their relationships between parts of a PDF document. 
\comment{
However, they are made up of common components, which take up a much smaller portion of the specification, and are used to store data and internal references. These components are rigidly structured, and there is great repetition in their use.

While the components take up a (relatively) small part of the specification, there are still many of them and they are tedious to encode by hand. Furthermore, their use in the specific data objects is varied and complex. This combination of data and text-encoded data structures which have some similarity but varied uses is what makes PDF data objects such an attractive target for learning, and at the same time such a challenge to learn.
}

PDF files are encoded in a textual format, which may contain binary information streams (e.g., images, encrypted data).
A PDF document is a sequence of at least one PDF body.
A PDF body is composed of three sections: objects, cross-reference table, and trailer.

\paragraph{Objects.}
The data and metadata in a PDF document is organized in basic units called objects. Objects are all similarly formatted, as seen in \Cref{pdf-samples}(a), and have a joint outer structure.
The first line of the object is its identifier, for indirect references, its generation number, which is incremented if the object is overridden with a newer version, and ``\scode{obj}'' which indicates the start of an object. The ``\scode{endobj}'' indicator closes the object.

The object in \Cref{pdf-samples}(a) contains a dictionary structure, which is delimited by ``\scode{<<}'' and ``\scode{>>}'', and contains keys that begin with \scode{/} followed by their values. \scode{[ 3 0 R ]} is a cross-object reference to an object in the same document with the identifier $3$ and the generation number $0$. Since a document can be very large, a referenced object is accessed using random-access via a cross-reference table.

\begin{figure}[t]
\centering
\newcolumntype{C}[1]{>{\centering\let\newline\\\arraybackslash\hspace{0pt}}m{#1}}
\begin{tabular}{cC{0.2in}cC{0.2in}c}
\begin{lstlisting}
125 0 obj
[680.6 680.6]
endobj
\end{lstlisting} & &
\begin{lstlisting}
88 0 obj
(Related Work)
endobj
\end{lstlisting} & &
\begin{lstlisting}
75 0 obj
4171
endobj
\end{lstlisting}\\
(a) & & (b) & & (c)
\end{tabular}
\begin{tabular}{c}
\\
\begin{lstlisting}
47 1 obj
[false 170 85.5 (Hello) /My#20Name]
endobj
\end{lstlisting}\\
(d)
\end{tabular}
\caption{PDF data objects of different types.}\label{object-samples}
\end{figure}

Other examples of objects are shown in \Cref{object-samples}. The object in \Cref{object-samples}(a) has the content \scode{[680.6 680.6]}, which is an \emph{array object}. Its purpose is to hold coordinates referenced by another object. \Cref{object-samples}(b) is a string literal that holds the bookmark text for a PDF document section. \Cref{object-samples}(c) is a numeric object. \Cref{object-samples}(d) is an object containing a multi-type array. These are all examples of object types that are both used on their own and as the basic blocks from which other objects are composed (e.g., the dictionary object in \Cref{pdf-samples}(a) contains an array). The rules for defining and composing objects comprises the majority of the PDF-format specification.

\paragraph{Cross reference table.}
The cross reference tables of a PDF body contain the address in bytes of referenced objects within the document. \Cref{pdf-samples}(b) shows a cross-reference table with a subsection that contains the addresses for five objects with identifiers $1$-$5$ and the placeholder for identifier $0$ which never refers to an object. The object being pointed to is determined by the row of the table (the subsection will include $6$ objects starting with identifier $0$) where \scode{n} is an indicator for an object in use, where the first column is the address of the object in the file, and \scode{f} is an object not used, where the first column refers to the identifier of the previous free object, or in the case of object $0$ to object $65535$, the last available object ID, closing the circle.

\paragraph{Trailer.}
The trailer of a PDF body contains a dictionary (again contained within ``\scode{<<}'' and ``\scode{>>}'') of information about the body, and \scode{startxref} which is the address of the cross-reference table. This allows the body to be parsed from the end, reading \scode{startxref}, then skipping back to the cross-reference table and parsing it, and only parsing objects as they are needed.

\paragraph{Updating a document.}
PDF documents can be {\em updated incrementally}. This means that if a PDF writer wishes to update the data in object $12$, it will start a new PDF body, in it write the new object with identifier $12$, and a generation number greater than the one that appeared before. It will then write a new cross-reference table pointing to the new object, and append this body to the previous document. Similarly, an object will be deleted by creating a new cross-reference table and marking it as free. We use this method in order to append new objects in a PDF file, as discussed later in Section~\ref{sec:evaluation}.

\paragraph{Scope of this work.}
In this paper, we investigate how to leverage and adapt
neural-network-based learning techniques to learn a grammar for {\em
non-binary PDF data objects}. Such data objects are formatted text,
such as shown in \Cref{pdf-samples}(a) and \Cref{object-samples}.
Rules for defining and composing such data objects makes the bulk of
the 1,300-pages PDF-format specification. These rules are numerous and
tedious, but repetitive and structured, and therefore well-suited for
learning with neural networks (as we will show later). In contrast,
learning automatically the structure (rules) for defining
cross-reference tables and trailers, which involve constraints on
lists, addresses, pointers and counters, look too complex and less
promising for learning with neural networks. We also do not consider
binary data objects, which are encoded in binary (e.g., image)
sub-formats and for which fully-automatic blackbox and whitebox
fuzzing are already effective.

\section{Statistical Learning of Object Contents}\label{sec:learning}

We now describe our statistical learning approach for learning a generative model of PDF objects. The main idea is to learn a generative language model over the set of PDF object characters given a large corpus of objects. We use a sequence-to-sequence (seq2seq)~\cite{seq2seq,machinetranslation} network model that has been shown to produce state-of-the-art results for many different learning tasks such as machine translation~\cite{machinetranslation} and speech recognition~\cite{speechrecognition}. The seq2seq model allows for learning arbitrary length contexts to predict next sequence of characters as compared to traditional n-gram based approaches that are limited by contexts of finite length. Given a corpus of PDF objects, the seq2seq model can be trained in an unsupervised manner to learn a generative model to generate new PDF objects using a set of input and output sequences. The input sequences correspond to sequences of characters in PDF objects and the corresponding output sequences are obtained by shifting the input sequences by one position. The learnt model can then be used to generate new sequences (PDF objects) by sampling the distribution given a starting prefix (such as \quotes{\texttt{obj}}).

\subsection{Sequence-to-Sequence Neural Network Models}

A recurrent neural network (RNN) is a neural network that operates on a variable length input sequence $\langle x_1,x_2,\cdots,x_T \rangle$ and consists of a hidden state $h$ and an output $y$. The RNN processes the input sequence in a series of time stamps (one for each element in the sequence). For a given time stamp $t$, the hidden state $h_t$ at that time stamp and the output $y_t$ is computed as:
\begin{equation*}
h_t = f(h_{t-1},x_{t})
\end{equation*}
\begin{equation*}
y_t = \phi(h_t)
\end{equation*}
where $f$ is a non-linear activation function such as sigmoid, $\tanh$ etc. and $\phi$ is a function such as \texttt{softmax} that computes the output probability distribution over a given vocabulary conditioned on the current hidden state. RNNs can learn a probability distribution over a character sequence $\langle x_1,\cdots,x_{t-1} \rangle$ by training to predict the next character $x_t$ in the sequence, i.e., it can learn the conditional distribution $p(x_t|\langle x_1,\cdots,x_{t-1} \rangle)$.

\begin{figure}[t]
\centering
\includegraphics[scale=0.4]{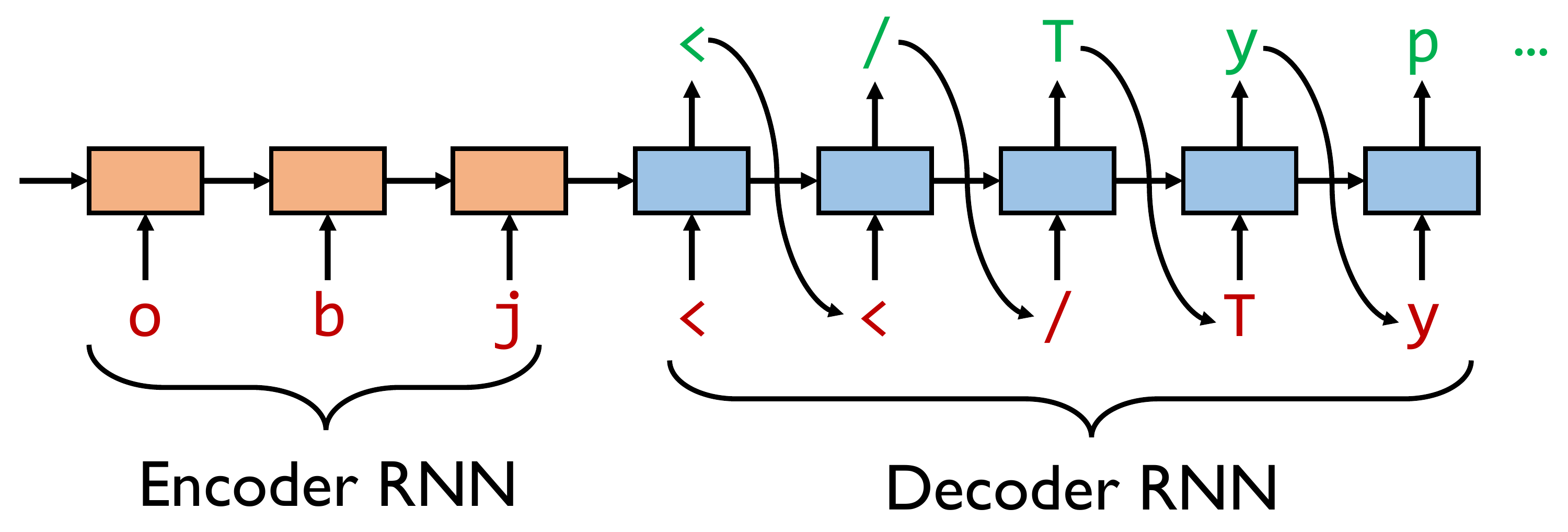}
\caption{A sequence-to-sequence RNN model to generate PDF objects.}
\label{seqseq}
\end{figure}

Cho et al.~\cite{seq2seq} introduced a sequence-to-sequence (seq2seq) model that consists of two recurrent neural networks, an encoder RNN that processes a variable dimensional input sequence to a fixed dimensional representation, and a decoder RNN that takes the fixed dimensional input sequence representation and generates the variable dimensional output sequence. The decoder network generates output sequences by using the predicted output character generated at time step $t$ as the input character for timestep $t+1$. An illustration of the \texttt{seq2seq} architecture is shown in Figure.~\ref{seqseq}. This architecture allows us to learn a conditional distribution over a sequence of next outputs, i.e., $p( \langle y_1,\cdots,y_{T_1} \rangle | \langle x_1,\cdots,x_{T_2} \rangle)$.

We train the seq2seq model using a corpus of PDF objects treating each one of them as a sequence of characters. During training, we first concatenate all the object files $s_i$ into a single file resulting in a large sequence of characters $\tilde{s} = s_1 + \cdots + s_n$. We then split the sequence into multiple training sequences of a fixed size $d$, such that the $i^{\texttt{th}}$ training instance $t_i = \tilde{s}[i*d:(i+1)*d]$, where $s[k:l]$ denotes the subsequence of $s$ between indices $k$ and $l$. The output sequence for each training sequence is the input sequence shifted by $1$ position, i.e., $o_t=\tilde{s}[i*d+1:(i+1)*d+1]$. The seq2seq model is then trained end-to-end to learn a generative model over the set of all training instances.

\subsection{Generating new PDF objects}

We use the learnt seq2seq model to generate new PDF objects. There are many different strategies for object generation depending upon the sampling strategy used to sample the learnt distribution. We always start with a prefix of the sequence \quotes{\texttt{obj }} (denoting the start of an object instance), and then query the model to generate a sequence of output characters until it produces \quotes{\texttt{endobj}} corresponding to the end of the object instance. We now describe three different sampling strategies we employ for generating new object instances.

\paragraph{{\bf \nosample}:} In this generation strategy, we use the learnt distribution to greedily predict the best character given a prefix. This strategy results in generating PDF objects that are most likely to be well-formed and consistent, but it also limits the number of objects that can be generated. Given a prefix like \quotes{\texttt{obj}}, the best sequence of next characters is uniquely determined and therefore this strategy results in the same PDF object. This limitation precludes this strategy from being useful for fuzzing.

\paragraph{{\bf \orig}:} In this generation strategy, we use the learnt distribution to \emph{sample} next characters (instead of selecting the top predicted character) in the sequence given a prefix sequence. This sampling strategy is able to generate a diverse set of new PDF objects by combining various patterns the model has learnt from the diverse set of objects in the training corpus. Because of sampling, the generated PDF objects are not always guaranteed to be well-formed, which is useful from the fuzzing perspective.

\paragraph{{\bf \fuzz}:} This sampling strategy is a combination of $\orig$ and $\nosample$ strategies. It samples the distribution to generate the next character only when the current prefix sequence ends with a whitespace, whereas it uses the best character from the distribution in middle of tokens (i.e., prefixes ending with non-whitespace characters), similar to the $\nosample$ strategy. This strategy is expected to generate more well-formed PDF objects compared to the $\orig$ strategy as the sampling is restricted to only at the end of whitespace characters.

\subsection{\textsc{SampleFuzz}: Sampling with Fuzzing}

Our goal of learning a generative model of PDF objects is ultimately to perform fuzzing. A perfect learning technique would always generate well-formed objects that would not exercise any error-hanlding code, whereas a bad learning technique would result in ill-formed objects that woult be quickly rejected by the parser upfront. To explore this tradeoff, we present a new algorithm, dubbed \t{SampleFuzz}, to perform some fuzzing while sampling new objects. We use the learnt model to generate new PDF object instances, but at the same time introduce anomalies to exercise error-handling code. 

The \t{SampleFuzz} algorithm is shown in Algorithm~\ref{samplefuzzalgo}. It takes as input the learnt distribution $\mathcal{D}(\t{x},\theta)$, the probability of fuzzing a character $t_\t{fuzz}$, and a threshold probability $p_t$ that is used to decide whether to modify the predicted character. While generating the output sequence \t{seq}, the algorithm samples the learnt model to get some next character $c$ and its probability $p(c)$ at a particular timestamp $t$. If the probability $p(c)$ is higher than a user-provided threshold $p_t$, i.e., if the model is confident that $c$ is likely the next character in the sequence, the algorithm chooses to instead sample another different character $c'$ in its place where $c'$ has the minimum probability $p(c')$ in the learnt distribution. This modification (fuzzing) takes place only if the result $p_\t{fuzz}$ of a random coin toss returns a probability higher than input parameter $t_\t{fuzz}$, which lets the user further control the probability of fuzzing characters. The key intuition of the \t{SampleFuzz} algorithm is to introduce unexpected characters in objects only in places where the model is {\em highly confident}, in order to trick the PDF parser. The algorithm also ensures that the object length is bounded by \t{MAXLEN}. Note that the algorithm is not guaranteed to always terminate, but we observe that it always terminates in practice.

\begin{algorithm}[t]
\caption{\t{SampleFuzz}($\mathcal{D}(\t{x},\theta),t_\t{fuzz}, p_t$)}
\begin{algorithmic}
\STATE {\t{seq} := \quotes{obj }}
\WHILE{$\neg$ \t{seq}.\t{endswith}(\quotes{endobj})}
\STATE{c,p(c) := \t{sample}($\mathcal{D}$(\t{seq},$\theta$))} (* Sample c from the learnt distribution *)
\STATE{$p_\t{fuzz} := \t{random}(0,1) $} (* random variable to decide whether to fuzz *)
\IF{$p_\t{fuzz} > t_\t{fuzz} \land p(c) > p_t$}
\STATE{c := $\argmin_{c'} \{p(c') \sim \mathcal{D}(\t{seq},\theta)\} $} (* replace c by c' (with lowest likelihood) *)
\ENDIF
\STATE{\t{seq} := \t{seq} + c}
\IF{\t{len(seq)} $>$ \t{MAXLEN}}
\STATE{\t{seq} := \quotes{obj }} (* Reset the sequence *)
\ENDIF
\ENDWHILE
\RETURN{\t{seq}}
\end{algorithmic}
\label{samplefuzzalgo}
\end{algorithm}

\subsection{Training the Model}

Since we train the seq2seq model in an unsupervised learning setting, we do not have test labels to explicitly determine how well the learnt models are performing. We instead train multiple models parameterized by number of passes, called \emph{epochs}, that the learning algorithm performs over the training dataset. An \emph{epoch} is thus defined as an iteration of the learning algorithm to go over the complete training dataset. We evaluate the seq2seq models trained for five different numbers of epochs: 10, 20, 30, 40, and 50. In our setting, one epoch takes about 12 minutes to train the seq2seq model, and the model with 50 epochs takes about 10 hours to learn. We use an LSTM model~\cite{lstm} (a variant of RNN) with 2 hidden layers, where each layer consists of 128 hidden states. 

\section{Experimental Evaluation}\label{sec:evaluation}

\subsection{Experiment Setup}

In this section, we present results of various fuzzing experiments
with the PDF viewer included in Microsoft's new Edge browser. We used
a self-contained single-process test-driver executable provided by
the Windows team for testing/fuzzing purposes. \comment{anonymized: provided
to us by the Windows organization.} This executable takes a PDF file
as input argument, executes the PDF parser included in the Microsoft
Edge browser, and then stops.  If the executable detects any parsing
error due to the PDF input file being malformed, it prints an error
message in an execution log.  In what follows, we simply refer to it
as the {\em Edge PDF parser}. All experiments were performed on 4-core
64-bit Windows 10 VMs with 20Gb of RAM.

We use three main standard metrics to measure fuzzing effectiveness:
\begin{description}
\topsep0pt
\itemsep0pt
\item [Coverage.] For each test execution, we measure instruction coverage, that is, the set of all unique instructions executed during that test. Each instruction is uniquely identified by a pair of values {\tt dll-name} and {\tt dll-offset}. The coverage for a set of tests is simply the union of the coverage sets of each individual test.

\item [Pass rate.] For each test execution, we programmatically check ({\tt grep}) for the presence of parsing-error messages in the PDF-parser execution log. If there are no error messages, we call this test {\em pass} otherwise we call it {\em fail}. Pass tests corresponds to PDF files that are considered to be well-formed by the Edge PDF parser. This metric is less important for fuzzing purposes, but it will help us estimate the quality of the learning.

\item [Bugs.] Each test execution is performed under the monitoring of the tool AppVerifier, a free runtime monitoring tool that can catch memory corruptions bugs (such as buffer overflows) with a low runtime overhead (typically a few percent runtime overhead) and that is widely used for fuzzing on Windows (for instance, this is how SAGE~\cite{SAGE} detects bugs).

\end{description}

\subsection{Training Data}

We extracted about 63,000 non-binary PDF objects out of a diverse set
of 534 PDF files. These 534 files themselves were
provided to us by the Windows fuzzing team and had been used for prior
extended fuzzing of the Edge PDF parser. This set of 534
files was itself the result of {\em seed minimization}, that is, the
process of computing a subset of a larger set of input files which
provides the same instruction coverage as the larger set. Seed
minimization is a standard first step applied before file
fuzzing~\cite{fuzzing-book,SAGE}. The larger set of PDF files came
from various sources, like past PDF files used for fuzzing but also
other PDF files collected from the public web. \comment{anonymized:
and our own intranet.}

These 63,000 non-binary objects are the training set for the RNNs we
used in this work. Binary objects embedded in PDF files (typically
representing images in various image formats) were not considered in
this work.

We learn, generate, and fuzz PDF objects, but the Edge PDF
parser processes full PDF files, not single objects. Therefore we wrote a simple
program to correctly {\em append} a new PDF object to an existing
(well-formed) PDF file, which we call a {\em host}, following the
procedure discussed in Section~\ref{pdf-struc} for updating a PDF
document. Specifically, this program first identifies the last trailer
in the PDF host file. This provides information about the file, such
as addresses of objects and the cross-reference table, and the last used object
ID. Next, a new body section is added to the file. In it, the new
object is included with an object ID that overrides the last object in
the host file. A new cross reference table is appended, which
increases the generation number of the overridden object. Finally, a
new trailer is appended.

\subsection{Baseline Coverage}

To allow for a meaningful interpretation of coverage results, we
randomly selected 1,000 PDF objects out of our 63,000 training
objects, and we measured their coverage of the Edge PDF parser, to be
used as a baseline for later experiments.

A first question is which host PDF file should we use in our
experiments: since any PDF file will have some objects in it, will a
new appended object interfere with other objects already present in
the host, and hence influence the overall coverage and pass rate?

To study this question, we selected the smallest three PDF files in
our set of 534 files, and used those as hosts. These three hosts are
of size 26Kb, 33Kb and 16Kb respectively.

\begin{figure}[t]
\centering
\includegraphics[scale=0.3]{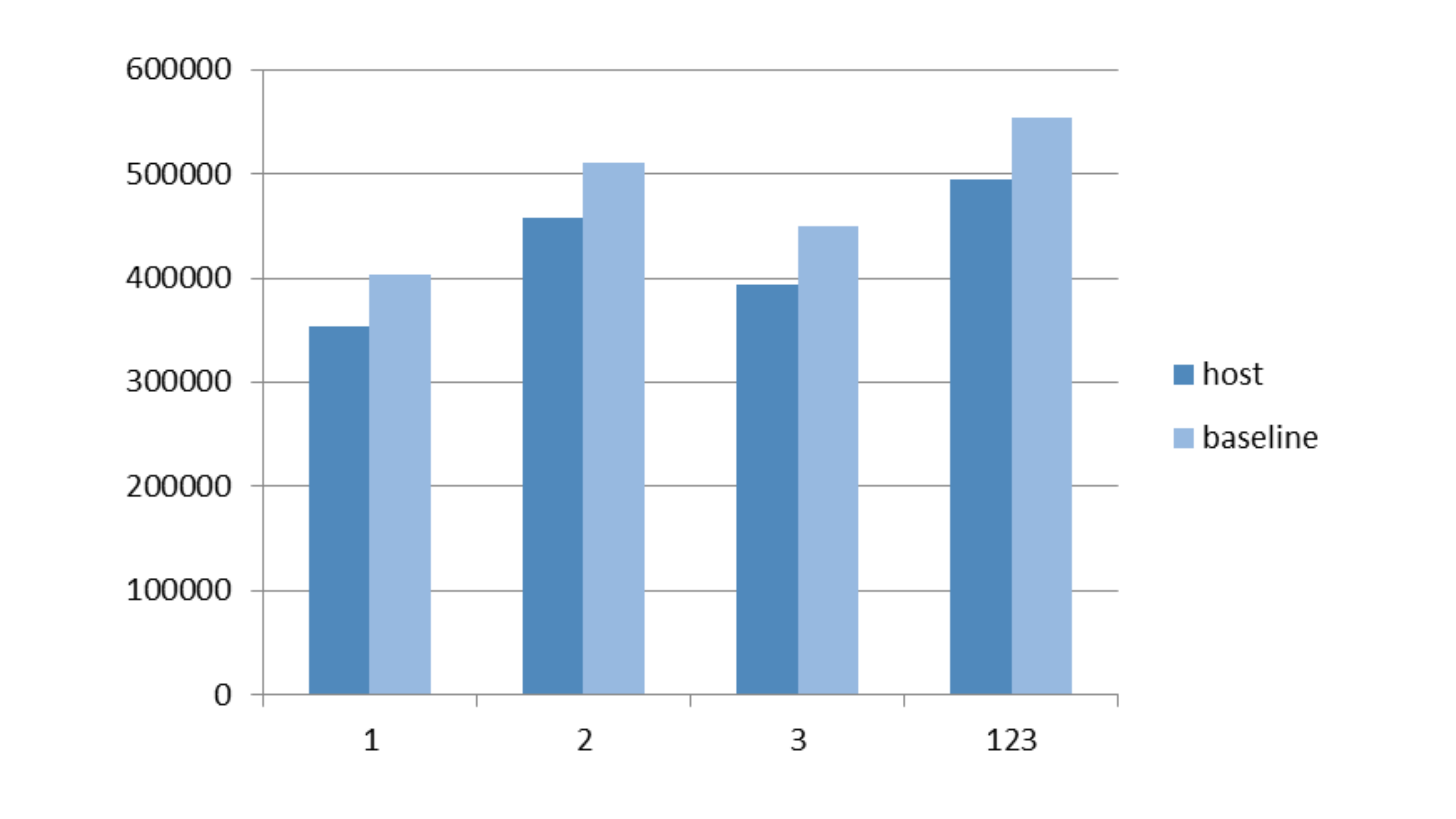}
\vspace*{-0.5cm}
\caption{Coverage for PDF hosts and baselines.}
\label{fig:baseline-coverage}
\end{figure}

Figure~\ref{fig:baseline-coverage} shows the instruction coverage
obtained by running the Edge PDF parser on the three hosts, denoted
{\tt host1}, {\tt host2}, and {\tt host3}. It also show the coverage
obtained by computing the union of these three sets, denoted {\tt
host123}. Coverage ranges from 353,327 ({\tt host1}) to 457,464 ({\tt
host2}) unique instructions, while the union ({\tt host123}) is 494,652
and larger than all three -- each host covers some unique instructions
not covered by the other two. Note that the smallest file {\tt host3}
does not lead to the smallest coverage.

Next, we recombined each of our 1,000 baseline objects with each of
our three hosts, to obtain three sets of 1,000 new PDF files, denoted
{\tt baseline1}, {\tt baseline2} and {\tt baseline3},
respectively. Figure~\ref{fig:baseline-coverage} shows the coverage of
each set, as well as their union {\tt baseline123}. We observe the
following.
\begin{itemize}
\topsep0pt
\itemsep0pt
\item The baseline coverage varies
depending on the host, but is larger than the host alone (as
expected). The largest difference between a host and a baseline
coverage is 59,221 instruction for {\tt host123} out of 553,873
instruction for {\tt baseline123}. In other words, 90\% of all
instructions are included in the host coverage no matter what new
objects are appended.

\item Each test typically covers on the
order of half a million unique instructions; this confirms that the
Edge PDF parser is a large and non-trivial application.

\item 1,000 PDF files take about 90 minutes to be processed (both to be
tested and get the coverage data).

\end{itemize}
We also measured the pass rate for each experiment. As expected, the
pass rate is 100\% for all 3 hosts.

{\bf Main Takeaway:} Even though coverage varies across hosts because
objects may interact differently with each host, the re-combined PDF
file is always perceived as well-formed by the Edge PDF parser.

\subsection{Learning PDF Objects}

When training the RNN, an important parameter is the number of epochs
being used (see Section~\ref{sec:learning}). We report here results of
experiments obtained after training the RNN for 10, 20, 30, 40, and 50
epochs, respectively. After training, we used each learnt RNN model to generate 1,000 unique PDF objects. We also compared the generated objects with the 63,000 objects used for training the model, and found no exact matches.

As explained earlier in Section~\ref{sec:learning}, we consider two
main RNN generation modes: the $\orig$ mode where we sample the
distribution at every character position, and the $\fuzz$ mode where we sample
the distribution only after whitespaces and generate the top predicted character for other positions.

\begin{figure}[t]
\centering
\includegraphics[scale=0.3]{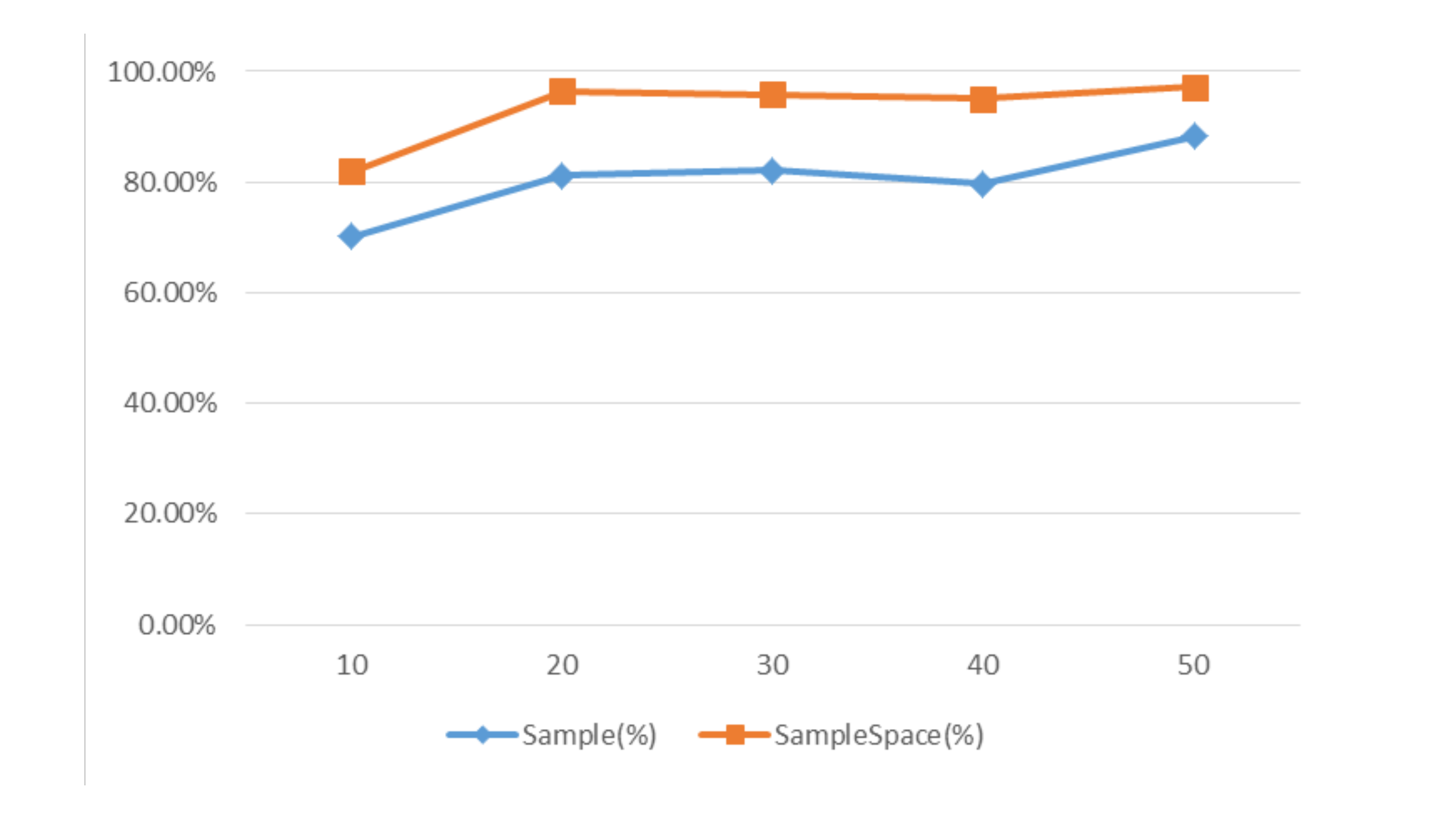}
\vspace*{-0.5cm}
\caption{Pass rate for $\orig$ and $\fuzz$ from 10 to 50 epochs.}
\label{fig:passRate}
\end{figure}

The pass rate for $\orig$ and $\fuzz$ when training with 10 to 50 epochs is
reported in Figure~\ref{fig:passRate}. We observe the following:
\begin{itemize}
\topsep0pt
\itemsep0pt
\item The pass rate for $\fuzz$ is consistently better than the one for $\orig$.
\item For 10 epochs only, the pass rate for $\orig$ is already above 70\%. This means that the learning is of good quality.
\item As the number of epochs increases, the pass rate increases, as expected, since the learned models become more precise but they also take more time (see Section~\ref{sec:learning}).
\item The best pass rate is 97\% obtained with $\fuzz$ and 50 epochs.
\end{itemize}
Interestingly, the pass rate is essentially the same regardless of the
host PDF file being used: it varies by at most 0.1\% across hosts (data not shown here).

{\bf Main Takeaway:} The pass rate ranges between $70\%$ and $97\%$
and shows the learning is of good quality.

\subsection{Coverage with Learned PDF Objects}

\begin{figure}[t]
\centering
\hspace*{-2cm}
\includegraphics[scale=0.5]{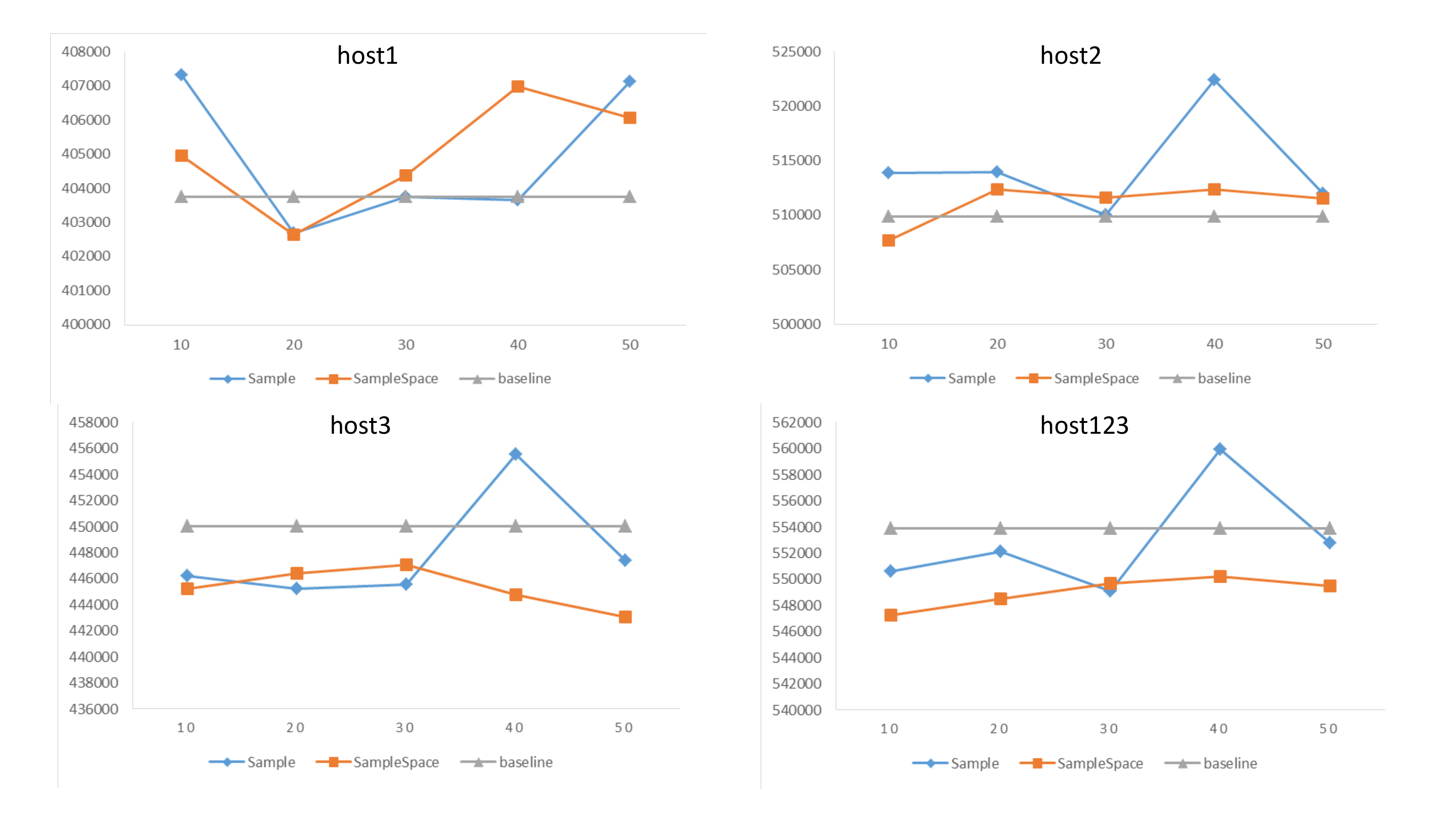}
\vspace*{-0.5cm}
\caption{Coverage for $\orig$ and $\fuzz$ from 10 to 50 epochs, for {\tt host 1, 2, 3,} and {\tt 123}.}
\label{fig:epochs-coverage}
\end{figure}

Figure~\ref{fig:epochs-coverage} shows the instruction coverage
obtained with $\orig$ and $\fuzz$ from 10 to 50 epochs and using {\tt host1}
(top left), {\tt host2} (top right), {\tt host3} (bottom left), and the
overall coverage for all hosts {\tt host123} (bottom right). The
figure also shows the coverage obtained with the corresponding {\tt
baseline}. We observe the following:
\begin{itemize}
\topsep0pt
\itemsep0pt
\item Unlike for the pass rate, the host impacts coverage significantly, as already pointed out earlier. Moreover, the shapes of each line vary across hosts.

\item For {\tt host1} and {\tt host2}, the coverage for $\orig$ and $\fuzz$ are above the {\tt baseline} coverage for most epoch results, while they are mostly below the {\tt baseline} coverage for {\tt host3} and {\tt host123}.

\item The best overall coverage is obtained with $\orig$ 40-epochs (see the {\tt host123} data at the bottom right).

\item The {\tt baseline123} coverage is overall second best behind $\orig$ 40-epochs.

\item The best coverage obtained with $\fuzz$ is also with 40-epochs.

\end{itemize}
{\bf Main Takeaway:} The best overall coverage is obtained with $\orig$ 40-epochs.

\subsection{Comparing Coverage Sets}

\begin{figure}[t]
\centering
\begin{tabular}{c|c|c|c|c}
Row$\setminus$Column & $\orig$-40e & $\fuzz$-40e & {\tt baseline123} & {\tt host123} \\
\hline
$\orig$-40e & 0 & 10,799 & 6,658 & 65,442 \\
$\fuzz$-40e & 1,680 & 0 & 3,393 & 56,323 \\
{\tt baseline123} & 660 & 6,514 & 0 & 59,444 \\
{\tt host123} & 188 & 781 & 223 & 0 \\
\end{tabular}
\caption{Comparing coverage: unique instructions in each row compared to each column.}
\label{fig:coverage-overlap}
\end{figure}

So far, we simply counted the number of unique instructions being
covered. We now drill down into the overall {\tt host123} coverage
data of Figure~\ref{fig:epochs-coverage}, and compute the overlap
between overall coverage sets obtained with our 40-epochs winner
$\orig$-40e and $\fuzz$-40e, as well as the {\tt baseline123} and {\tt
host123} overall coverage. The results are presented in
Figure~\ref{fig:coverage-overlap}. We observe the following:
\begin{itemize}
\topsep0pt
\itemsep0pt
\item All sets are almost supersets of {\tt host123} as expected (see the {\tt host123} row), except for a few hundred instructions each.

\item $\orig$-40e is almost a superset of all other sets,
except for 1,680 instructions compared to $\fuzz$-40e, and a few
hundreds instructions compared to {\tt baseline123} and {\tt host123}
(see the $\orig$-40e column).

\item $\orig$-40e and $\fuzz$-40e have way more instructions in common
than they differ (10,799 and 1,680), with $\orig$-40e having better
coverage than $\fuzz$-40e.

\item $\fuzz$-40e is incomparable with {\tt baseline123}: it has 3,393 more instructions but also 6,514 missing instructions.

\end{itemize}
{\bf Main Takeaway:} Our coverage winner $\orig$-40e is almost a
superset of all other coverage sets.

\subsection{Combining Learning and Fuzzing}

In this section, we consider several ways to combine learning with
fuzzing, and evaluate their effectiveness.

We consider a widely-used simple blackbox random fuzzing algorithm,
denoted {\tt Random}, which randomly picks a position in a file and
then replaces the byte value by a random value between 0 and 255. The
algorithm uses a {\em fuzz-factor} of 100: the length of the file
divided by 100 is the average number of bytes that are fuzzed in that
file.

We use {\tt random} to generate 10 variants of every PDF object
generated by 40-epochs $\orig$-40e, $\fuzz$-40e, and {\tt
baseline}. The resulting fuzzed objects are re-combined with our 3
host files, to obtain three sets of 30,000 new PDF files, denoted by
$\origrandom$, $\fuzzrandom$ and {\tt baseline+Random}, respectively.

For comparison purposes, we also include the results of running
$\orig$-40e to generate 10,000 objects, denoted $\orig$-10K.

Finally, we consider our new algorithm $\morefuzz$ described in
Section~\ref{sec:learning}, which decides where to fuzz values based on the
learnt distribution. We applied this algorithm with the learnt
distribution of the 40-epochs RNN model, $t_\t{fuzz} = 0.9$,
and a threshold $p_t = 0.9$.

\begin{figure}[t]
\centering
\begin{tabular}{c|c|c}
Algorithm & Coverage & Pass Rate \\
\hline
$\fuzzrandom$ & 563,930 &         36.97\%\\
{\tt baseline+Random} & 564,195 & 44.05\%\\
$\orig$-10K & 565,590 &           78.92\% \\
$\origrandom$ & 566,964 &         41.81\%\\
$\morefuzz$ &  567,634 &          68.24\% \\
\end{tabular}
\caption{Results of fuzzing experiments with 30,000 PDF files each.}
\label{fig:fuzzing-results}
\end{figure}

Figure~\ref{fig:fuzzing-results} reports the overall coverage and the
pass-rate for each set. Each set of 30,000 PDF files takes about 45
hours to be processed. The rows are sorted by increasing coverage.
We observe the following:
\begin{itemize}
\topsep0pt
\itemsep0pt
\item After applying {\tt Random} on objects generated with $\orig$, $\fuzz$ and {\tt baseline}, coverage goes up while the pass rate goes down: it is consistently below $50\%$.

\item After analyzing the overlap among coverage sets (data not shown here), all fuzzed sets are almost supersets of their original non-fuzzed sets (as expected).

\item Coverage for $\orig$-10K also increases by 6,173 instructions compared to $\orig$, while the pass rate remains around $80\%$ (as expected).

\item Perhaps surprisingly, the best overall coverage is obtained with $\morefuzz$. Its pass rate is $68.24\%$.

\item The difference in absolute coverage between $\morefuzz$ and the next best $\origrandom$ is only 670 instructions. Moreover, after analyzing the coverage set overlap, $\morefuzz$ covers 2,622 more instructions than $\origrandom$, but also misses 1,952 instructions covered by $\origrandom$. Therefore, none of these two top-coverage winners fully ``simulate'' the effects of the other.
\end{itemize}
{\bf Main Takeaway:} All the learning-based algorithms considered here
are competitive compared to {\tt baseline+Random}, and three of those
beat that baseline coverage.

\subsection{Main Takeaway: Tension between Coverage and Pass Rate}

The main takeaway from all our experiments is the {\em tension we
observe between the coverage and the pass rate}.

This tension is visible in Figure~\ref{fig:fuzzing-results}. But it is
also visible in earlier results: if we correlate the coverage results
of Figure~\ref{fig:epochs-coverage} with the pass-rate results of
Figure~\ref{fig:passRate}, we can clearly see that $\fuzz$ has a
better pass rate than $\orig$, but $\orig$ has a better overall
coverage than $\fuzz$ (see {\tt host123} in the bottom right of
Figure~\ref{fig:epochs-coverage}).

Intuitively, this tension can be explained as follows.  A pure
learning algorithm with a nearly-perfect pass-rate (like $\fuzz$)
generates almost only well-formed objects and exercises little
error-handling code.  In contrast, a {\em noisier} learning algorithm
(like $\orig$) with a lower pass-rate can not only generate many
well-formed objects, but it also generates some ill-formed ones which
exercise error-handling code.

Applying a random fuzzing algorithm (like {\tt random}) to
previously-generated (nearly) well-formed objects has an even more
dramatic effect on lowering the pass rate (see
Figure~\ref{fig:fuzzing-results}) while increasing coverage, again
probably due to increased coverage of error-handling code.

The new $\morefuzz$ algorithm seems to hit a sweet spot between both
pass rate and coverage. In our experiments, the sweet spot for the
pass rate seems to be around $65\%-70\%$: {\em this pass rate is high
enough to generate diverse well-formed objects that cover a lot of
code in the PDF parser, yet low enough to also exercise error-handling
code in many parts of that parser.}

Note that instruction coverage is ultimately a better indicator of
fuzzing effectiveness than the pass rate, which is instead a
learning-quality metric.

\subsection{Bugs}

In addition to coverage and pass rate, a third metric of interest is
of course the number of bugs found. During the experiments previously
reported in this section, no bugs were found. Note that the Edge PDF
parser had been thoroughly fuzzed for months with other fuzzers
(including SAGE~\cite{SAGE}) before we performed
this study, and that all the bugs found during this prior fuzzing had
been fixed in the version of the PDF parser we used for this study.

However, during a longer experiment with $\origrandom$, 100,000
objects and 300,000 PDF files (which took nearly 5 days), a
stack-overflow bug was found in the Edge PDF parser: a regular-size
PDF file is generated (its size is 33Kb) but it triggers an unexpected
recursion in the parser, which ultimately results in a stack overflow.
This bug was later confirmed and fixed by the Microsoft Edge
development team. We plan to conduct other longer experiments in the
near future.

\section{Related Work}\label{sec:related-work}

\newcommand{\footnoteurl}[1]{\footnote{\scriptsize\url{#1}}}

\paragraph{Grammar-based fuzzing.}
Most popular blackbox random fuzzers today support some form of
grammar representation, e.g.,
Peach\footnoteurl{http://www.peachfuzzer.com/} and
SPIKE\footnoteurl{http://resources.infosecinstitute.com/fuzzer-automation-with-spike/},
among many others~\cite{fuzzing-book}. Work on grammar-based test
input generation started in
the~1970's~\cite{hanford1970agt,purdom1972sgt} and is related to
model-based testing~\cite{utting2006tmb}. Test generation from a
grammar is usually either
random~\cite{maurer1990gtd,sirer1999upg,coppit2005yeu} or
exaustive~\cite{lammel2006ccc}. Imperative
generation~\cite{quickcheck,BrettDGM07} is a related approach in which
a custom-made program generates the inputs (in effect, the program
encodes the grammar). Grammar-based fuzzing can also be combined with
whitebox fuzzing~\cite{MX07,GKL08}.

\paragraph{Learning grammars for grammar-based fuzzing.} Bastani et al.~\cite{bastani} present an algorithm to synthesize a context-free grammar given a set of input examples, which is then used to generate new inputs for fuzzing. This algorithm uses a set of generalization steps by introducing repetition and alternation constructs for regular expressions, and merging non-terminals for context-free grammars, which in turn results in a monotonic generalization of the input language. This technique is able to capture hierarchical properties of input formats, but is not well suited for formats such as PDF objects, which are relatively flat but include a large diverse set of content types and key-value pairs. Instead, our approach uses sequence-to-sequence neural-network models to learn {\em statistical} generative models of such flat formats. Moreover, learning a statistical model also allows for guiding additional fuzzing of the generated inputs.

AUTOGRAM~\cite{autogram} also learns (non-probabilistic) context-free grammars given a set of inputs but by dynamically observing how inputs are processed in a program. It instruments the program under test with dynamic taints that tags memory with input fragments they come from. The parts of the inputs that are processed by the program become syntactic entities in the grammar. Tupni~\cite{tupni} is another system that reverse engineers an input format from examples using a taint tracking mechanism that associate data structures with addresses in the application address space. Unlike our approach that treats the program under test as a black-box, AUTOGRAM and Tupni require access to the program for adding instrumentation, are more complex, and their applicability and precision for complex formats such as PDF objects is unclear.

\paragraph{Neural-networks-based program analysis.} There has been a lot of recent interest in using neural networks for program analysis and synthesis. Several neural architectures have been proposed to learn simple algorithms such as array sorting and copying~\cite{neuralram,neuralpi}. Neural FlashFill~\cite{neuralflashfill} uses novel neural architectures for encoding input-output examples and generating regular-expression-based programs in a domain specific language. Several seq2seq based models have been developed for learning to repair syntax errors in programs~\cite{synfix,deepfix,evanmooc}. These techniques learn a seq2seq model over a set of correct programs, and then use the learnt model to predict syntax corrections for buggy programs. Other related work optimizes assembly programs using neural representations~\cite{neuraloptimize}. In this paper, we present a novel application of seq2seq models to learn grammars from sample inputs for fuzzing purposes.

\section{Conclusion and Future Work}\label{sec:conclusion}

Grammar-based fuzzing is effective for fuzzing applications with
complex structured inputs provided a comprehensive input grammar is
available. This paper describes the first attempt at using
neural-network-based statistical learning techniques to automatically
generate input grammars from sample inputs. We presented and evaluated
algorithms that leverage recent advances in sequence learning by
neural networks, namely \t{seq2seq} recurrent neural networks, to
automatically learn a generative model of PDF objects. We devised
several sampling techniques to generate new PDF objects from the
learnt distribution. We show that the learnt models are not only
able to generate a large set of new well-formed objects, but also
results in increased coverage of the PDF parser used in our
experiments, compared to various forms of random fuzzing. 

While the results presented in Section~\ref{sec:evaluation} may vary
for other applications, our general observations about the tension
between conflicting learning and fuzzing goals will remain valid:
learning wants to capture the structure of well-formed inputs, while
fuzzing wants to break that structure in order to cover unexpected
code paths and find bugs. We believe that the inherent statistical
nature of learning by neural networks is a powerful tool to address
this learn\&fuzz challenge.

There are several interesting directions for future work. While the focus of our paper was on learning the structure of PDF objects, it would be worth exploring how to learn, as automatically as possible, the higher-level hierarchical structure of PDF documents involving cross-reference tables, object bodies, and trailer sections that maintain certain complex invariants amongst them. Perhaps some combination of logical inference techniques with neural networks could be powerful enough to achieve this. Also, our learning algorithm is currently agnostic to the application under test. We are considering using some form of reinforcement learning to guide the learning of \t{seq2seq} models with coverage feedback from the application, which could potentially guide the learning more explicitly towards increasing coverage.

{\bf Acknowledgments.} We thank Dustin Duran and Mark Wodrich from the Microsoft Windows security team for their Edge-PDF-parser test-driver and for helpful feedback. We also thank the team members of Project Springfield, which partly funded this work.

\bibliography{biblio}
\bibliographystyle{plain}

\end{document}